# Conditioned Generative Modeling of Molecular Glues: A Realistic AI Approach for Synthesizable Drug-Like Molecules

Naeyma N. Islam, PhD[1] and Thomas R. Caulfield, PhD [2,*]

[1] Department of Neuroscience, Mayo Clinic, Jacksonville FL; islam.naeyma@mayo.edu
[2] Digital Ether Computing, Miami, FL; thomas@digitalethercomputing.com
* Correspondence: thomas@digitalethercomputing.com;

**Abstract:** Alzheimer's disease (AD) is marked by the pathological accumulation of amyloid beta-42 (Aβ42), contributing to synaptic dysfunction and neurodegeneration. While extracellular amyloid plaques are well-studied, increasing evidence highlights intracellular Aβ42 as an early and toxic driver of disease progression. In this study, we present a novel, AI-assisted drug design approach to promote targeted degradation of Aβ42 via the ubiquitin-proteasome system (UPS), using E3 ligase-directed molecular glues. We systematically evaluated the ternary complex formation potential of Aβ42 with three E3 ligases-CRBN, VHL, and MDM2-through structure-based modeling, ADMET screening, and docking. We then developed a Ligase-Conditioned Junction Tree Variational Autoencoder (LC-JT-VAE) to generate ligase-specific small molecules, incorporating protein sequence embeddings and torsional angle-aware molecular graphs. Our results demonstrate that this generative model can produce chemically valid, novel, and target-specific molecular glues capable of facilitating Aβ42 degradation. This integrated approach offers a promising framework for designing UPS-targeted therapies for neurodegenerative diseases.

## 1. Introduction

Alzheimer's disease (AD), the most prevalent neurodegenerative disorder, is characterized by progressive cognitive decline and neuronal loss. The accumulation of amyloid beta 42 (Aβ42) peptides, derived from the aberrant cleavage of amyloid precursor protein (APP) by β- and γ-secretases, is a central pathological hallmark [1-4]. Aβ42 is particularly aggregation-prone and toxic compared to Aβ40, forming intracellular deposits that precede extracellular amyloid plaques and correlate with early synaptic dysfunction and neuronal death [1,5-8]. Thus, targeting intracellular Aβ42 is crucial for developing effective AD therapies.

Despite numerous therapeutic efforts-including secretase inhibition, immunotherapy, and enhancing proteostasis-significant challenges persist. Secretase inhibitors suffer from off-target toxicities due to the broad substrate specificity of γ- and β-secretases [9-11]. Immunotherapeutic strategies, while promising, face issues with blood-brain barrier (BBB) penetration and potential

CNS inflammation [12,13]. Enhancement of autophagy to clear Aβ42 aggregates has shown potential but remains nonspecific and may compromise essential cellular functions [14].

The ubiquitin-proteasome system (UPS) offers an attractive alternative for degrading intracellular Aβ42 aggregates [15,16]. However, Aβ42's aggregation-prone nature impedes its efficient recognition by the UPS. Molecular glues-small molecules that induce or stabilize interactions between E3 ligases and target proteins-have emerged as a promising strategy to facilitate Aβ42 degradation through ubiquitination and proteasomal clearance [15-19]. Unlike PROTACs, molecular glues require no bifunctional linker and can enhance BBB permeability due to their smaller size [17-19].

Selecting the appropriate E3 ligase is critical. VHL, CRBN, and MDM2 are prominent candidates based on their structural properties and prior success in targeted protein degradation [20-24]. VHL, associated with hypoxia signaling; CRBN, known for ligand-dependent substrate recruitment; and MDM2, a versatile p53 regulator, offer complementary substrate specificities that could be leveraged for Aβ42 targeting. In this work, we evaluate these three E3 ligases for their potential to form stable ternary complexes with Aβ42 and designed molecular glues.

Traditional molecular glue discovery relies on extensive chemical library screening, making it slow and resource-intensive. Artificial intelligence (AI) models now provide a powerful alternative, enabling efficient de novo design, lead optimization, and virtual screening [25-30]. Several generative AI models have been employed for small molecule design; however, each presents notable limitations. Sequence-based models, such as those relying on SMILES representations, often suffer from non-uniqueness and syntactic errors that lead to chemically invalid molecules, complicating the learning process and limiting their reliability in drug discovery [35]. Similarly, standard Variational Autoencoders (VAEs) and Generative Adversarial Networks (GANs), although effective at sampling chemical space, often fail to guarantee chemical validity and structural coherence in the generated compounds [36,37].

To overcome these challenges, we selected the Junction Tree Variational Autoencoder (JTVAE) [31-34], a graph-based model that encodes molecular structures as hierarchical trees of chemically meaningful substructures, enabling the generation of valid, drug-like molecules with superior structural fidelity. Unlike sequence-based or flat graph models, JTVAE preserves chemical rules during molecule generation, making it particularly well-suited for drug design applications. While the original JTVAE framework models molecular structures as 2D graphs, it does not account for torsional flexibility, a critical factor for molecular stability, bioactivity, and ligand-receptor interactions. To address this limitation, we extended JTVAE by explicitly integrating torsional angles into bond representations, enabling the model to capture conformational dynamics and generate structurally diverse, chemically realistic molecules with improved 3D predictive fidelity.Moreover, to design molecular glues specific to different E3 ligases, we further enhanced the model by incorporating protein-level information. Specifically, we embedded the amino acid sequences of binding site residues from CRBN, VHL, and MDM2 into the training pipeline. This allowed the model to learn ligase-specific binding patterns, enabling conditional generation of molecular glues optimized for forming stable ternary complexes between Aβ42 and individual E3 ligases. Together, these advancements substantially improve the chemical validity, conformational realism, and biological specificity of the molecules generated compared to the original JTVAE framework.

In this study, we investigate the potential of three widely studied E3 ubiquitin ligases-VHL, CRBN, and MDM2-for mediating the targeted degradation of amyloid beta (Aβ42) through molecular glue strategies. Furthermore, we design ligase-specific novel molecular glues by incorporating protein sequence embeddings derived from the binding site residues of each E3 ligase into a Junction Tree Variational Autoencoder (JT-VAE), a graph-based generative model. To enhance the precision of molecular glue generation, we also integrate three-dimensional conformational features-including molecular rotational angles-alongside conventional structural and chemical descriptors. This multi-modal approach allows for the generation of chemically valid, drug-like small molecules that are optimized for selective engagement with E3 ligases and effective degradation of Aβ42.

## 2. METHODS
### 2.1 Amyloid beta 42 structure

For structural studies of Aβ42, which corresponds to residues 627-713 of the full-length APP, we selected the cryo-EM structure of Type I Aβ42 filaments from human brain tissue (PDB ID: 7Q4B) [42]. These fibrils, resolved at 2.5 Å, represent a homo-decameric assembly of Aβ42 peptides adopting an S-shaped conformation, a structural feature crucial to their aggregation and stability in Alzheimer's disease pathology [43]. Although 7Q4B depicts extracellular fibrils, they originate from intracellular oligomerization processes, making this structure highly relevant to the pathological cascade [44].

The raw PDB structure was preprocessed using Schrödinger's Protein Preparation Wizard in Maestro [45]. Preprocessing steps included bond order assignment, addition of missing hydrogen atoms, removal of water molecules beyond 5 Å, protonation state adjustment with Epik at pH 7.0 ± 2.0, and optimization of the hydrogen bonding network. A restrained energy minimization was performed with the OPLS4 force field, allowing hydrogen atoms to relax while constraining heavy atoms. The resulting prepared structure was subsequently used for all in silico analyses.

### 2.2 Binding site identified in Amyloid Beta42

To identify ligandable regions on the Aβ42 fibril structure, we employed the SiteMap module within Schrödinger Maestro to predict and characterize potential binding pockets based on physicochemical and geometric criteria [45].

### 2.3 E3 Ligase Structure Selection and Preparation

High-quality crystallographic structures of three E3 ligases-VHL (PDB ID: 4W9C), CRBN (PDB ID: 4CI3), and MDM2 (PDB ID: 4OAS)-were selected based on resolution and relevance to small-molecule recruitment. The VHL structure (1.45 Å) offers a well-defined ligand-binding interface suitable for docking studies targeting Aβ42 degradation. The CRBN structure, resolved with thalidomide, reveals the canonical molecular glue binding site, while the MDM2 structure features a co-crystallized inhibitor occupying the druggable p53-interacting cleft. All structures were prepared using Schrödinger's Protein Preparation Wizard [45], with preprocessing steps including removal of non-essential heteroatoms and waters, bond order assignment, hydrogen addition, hydrogen bonding optimization, and restrained energy minimization using the OPLS3e or OPLS4 force fields. These curated models served as the structural foundation for E3 ligase-specific molecular glue design. Prepared structures of VHL, CRBN, and MDM2 are shown in **Figure 1**.

**2.4 ADMET based filtering and docking**

To optimize drug-likeness and safety, in silico ADMET (Absorption, Distribution, Metabolism, Excretion, and Toxicity) screening was performed on compound libraries from the Vitas and ChEMBL databases [46,47] using Schrödinger's QikProp and ADMET/Tox modules [48]. Compounds were filtered based on Lipinski's Rule of Five and Veber's rule, followed by predictions for intestinal absorption, blood-brain barrier permeability, CYP450 inhibition, hepatotoxicity, and AMES mutagenicity. Only candidates meeting all pharmacokinetic and toxicity thresholds were retained for downstream studies. Filtered compounds were docked into the ligand-binding sites of VHL, CRBN, and MDM2 using Schrödinger's Glide module in standard precision (SP) and extra precision (XP) modes. For each docked complex, Aβ42 was aligned near the E3-ligase interface, and ternary complexes were refined using Rosetta FlexPepDock and evaluated based on docking scores.

Additionally, all-atom molecular dynamics (MD) simulations were conducted using the Desmond simulation engine (Schrödinger Suite) with the OPLS4 force field. Ternary complexes were solvated in TIP3P water, neutralized with counterions, minimized, equilibrated under NPT conditions (300 K, 1 atm) using the Berendsen thermostat and barostat, and subjected to 100 ns production runs.

**2.5 Dataset Construction and Feature Engineering for Ligase-Conditioned Molecular Glue Design**

The training dataset consisted of ADMET-screened and docking-prioritized compounds capable of forming ternary complexes with CRBN, VHL, and MDM2 E3 ligases and amyloid-β42. All molecules were converted into molecular graph and junction tree formats using RDKit and the original JT-VAE preprocessing pipeline located at https://github.com/croningp/JTNN-VAE, which subdirectories for fast_molvae and python script preprocess.py at https://github.com/croningp/JTNN-VAE/blob/master/fast_molvae/preprocess.py. To enhance molecular representation beyond the capabilities of the original JT-VAE framework, we extended the preprocessing pipeline to incorporate torsional (dihedral) angle features for rotatable bonds. These dihedral angles were computed using a custom RDKit-based routine, which identifies rotatable bonds and calculates their dihedral values via the custom Python function "get_torsional_angle" (**Supplementary Figure S1**). This addition enables the model to better capture conformational flexibility in small molecules. Furthermore, each compound was paired with the corresponding E3 ligase's binding site sequence embedding, allowing the model to learn protein-contextual chemical features. Together, these enhancements-dihedral angle integration and protein sequence conditioning-enabled the development of a modified JT-VAE framework capable of generating E3 ligase-specific molecular glues for CRBN, VHL, and MDM2.

## 2.6 Development of Ligase-Conditioned JT-VAE Incorporating Torsional Features for Molecular Glue Design

To enable the generation of E3 ligase-specific molecular glues, we extended the canonical Junction Tree Variational Autoencoder (JT-VAE) framework introduced by Jin et al. [49], originally designed for chemically valid molecule generation without target protein context [50]. While the standard JT-VAE effectively encodes molecular graphs and junction trees, it lacks two critical capabilities for structure-guided drug design: (i) conditioning molecule generation on protein-specific information, and (ii) modeling three-dimensional (3D) conformational flexibility via torsional (dihedral) angles.

To address these limitations, we developed a Ligase-Conditioned JT-VAE (LC-JT-VAE). First, we enhanced the molecular graph representation by incorporating torsional angle features into the bond-level message passing steps. Dihedral angles were computed using RDKit's GetDihedralDeg function by identifying rotatable bonds (**Supplementary Figure S1**), providing critical 3D information on conformational dynamics essential for accurate ligand design. Second, we integrated sequence embeddings derived from the binding site residues of CRBN, VHL, and MDM2 directly into the molecular latent space, enabling the model to generate ligase-specific molecular glues.

## 2.7 Encoding E3 Ligase Binding Site Sequences

Binding site residue sequences for CRBN, VHL, and MDM2 were encoded through multiple strategies to capture structural and functional contexts. Initially, one-hot encoding and k-mer encoding were applied to extract local sequence patterns. For more advanced, context-aware representations, ProtBERT-a transformer model pretrained on UniRef100-was utilized to generate high-dimensional embeddings [51]. These embeddings were further refined using bidirectional long short-term memory (BiLSTM) networks [52] to model sequential dependencies, and then projected to fixed-length vectors (e.g., 128D or 512D) for compatibility with the molecular latent space, thus enabling protein-conditioned molecule generation.

## 2.8 LC-JT-VAE Architecture and Latent Space Fusion

The LC-JT-VAE architecture consists of two parallel encoders: a molecular graph encoder, comprising a message-passing neural network (MPN) and a junction tree message-passing network (JTMPN), and a protein sequence encoder for E3 ligase binding site embeddings. These two streams were fused in the latent space using either concatenation with linear projection and ReLU activation, or cross-attention mechanisms where the molecular graph attended to the protein embedding. The fused latent vector was passed to a conditional Junction Tree Neural Network (JTNN) decoder, which reconstructed the junction tree scaffold and molecular graph conditioned on the ligase binding context (**Figure 2**).

Implementation summary:

```
# During encoding
mol_latent = JTNNEncoder(mol_graph, junction_tree)
ligase_embed = LigaseEncoder(sequence)

# Latent fusion
z = torch.cat([mol_latent, ligase_embed], dim=1)

# During decoding
output = JTNNDecoder(z)
```

**2.9 Detailed Architectural Breakdown and Conditioning Mechanism**

- **Molecular Graph and Junction Tree Encoding**: Molecules were processed as graphs and junction trees using MPN and JTMPN. Bond features were enhanced by incorporating torsional angles computed via RDKit, improving the model's ability to capture conformational flexibility crucial for protein-ligand interactions.
- **Ligase Sequence Encoding**: Binding site sequences were embedded using ProtBERT or BiLSTM-based encoders and projected to match the molecular latent space dimensions.
- **Latent Space Fusion Strategies**:
    - Concatenation and Linear Projection:
      z_fused = ReLU(W[z_mol; z_seq] + b)
    - Cross-Attention Mechanism:
      z_mol' = MultiHeadAttention(z_mol, z_seq, z_seq)
- **Conditional Decoding**:
    - Tree Decoder: Reconstructs the junction tree scaffold.
    - Graph Decoder: Rebuilds the full molecular graph conditioned on the fused latent vector.

The detailed internal architecture of LC-JT-VAE, including the JTNNEncoder, GraphGRU layers, sequence encoder modules (W_H1, W_H2, W_O), and latent projections (T_mean, T_var, G_mean, G_var), is illustrated in **Supplementary Figure S2**. Together, these innovations distinguish our LC-JT-VAE from the original JT-VAE by enabling target-specific, conformation-aware molecular glue design.

**2.10 Model Training, Optimization, and Implementation Details**

The Ligase-Conditioned Junction Tree Variational Autoencoder (LC-JT-VAE) was trained using a *β*-VAE loss function, consisting of a reconstruction loss and a Kullback-Leibler (KL) divergence term to regularize the latent space [53,54]. The total loss was defined as:

$\mathcal{L}_{total} = \mathcal{L}_{recon} + \beta \cdot \mathcal{L}_{KL}$, where $\mathcal{L}_{total}$ included both molecular graph and junction tree reconstruction losses, and $\mathcal{L}_{KL}$ promoted latent space continuity. The annealing coefficient $\beta$ was initialized at 0 and incrementally increased by 0.002 per step until reaching 1.0. Model parameters were initialized with Xavier normal initialization for weights and constant initialization for biases. Training was performed over 2000 epochs using the Adam optimizer (learning rate = $1 \times 10^{-31}$), with exponential decay (factor = 0.9 per step) applied through an Exponential LR scheduler. Gradient norms were clipped at a maximum of 50.0 to prevent explosion during backpropagation. Mini-batches were generated using a custom tree-structured molecule loader. For each batch, the model computed the total loss, KL divergence, and three reconstruction metrics: word accuracy (wacc), topology accuracy (tacc), and stereochemistry accuracy (sacc), which were monitored across epochs for convergence evaluation. Model checkpoints were saved at each step for reproducibility and future fine-tuning.

All training, validation, and generation procedures were implemented in Python using PyTorch (v1.12) and executed on Google Colab Pro with GPU acceleration [55], utilizing NVIDIA L4 GPUs (22,000+ MiB VRAM), CUDA version 12.4, and driver version 550.54.15.

## 3. RESULTS

### 3.1 Identification of a Ligandable Interface on Amyloid-β42 Fibrils

The binding site (BS) analysis revealed three potential binding pockets on the fibrillar assembly as shown in the **Figure 3**. Of these, two sites were located in clefts (**Figure 3a-b**) between adjacent protofilaments. While such interfacial regions may appear druggable, their deep burial and steric shielding within the fibril core pose significant challenges for recruiting E3 ligases via bifunctional molecules or molecular glues. These buried sites are unlikely to support the geometric and spatial constraints required for ternary complex formation necessary for targeted protein degradation.

Instead, we selected the third site (**Figure 3c**), which lies at a more solvent-accessible interfacial region of the fibril surface. This site resides at a junction where structural accessibility and spatial exposure are compatible with simultaneous engagement of a small molecule and an E3 ligase receptor. Given the accessibility of this pocket and its favorable physicochemical enclosure, we prioritized it as the ligandable hotspot for downstream compound docking. **Figure 3d** illustrates the atomic representation of the binding site residues on Aβ42.

### 3.2 ADMET-Based Filtering of Compound Libraries

To prioritize drug-like compounds suitable for E3 ligase engagement and generative modeling, we applied a series of physicochemical and pharmacokinetic filters to the combined Vitas and ChEMBL compound libraries. The filtering criteria were based on predicted ADMET properties and included: molecular weight (MW) between 130 and 725 Da, lipophilicity (logPo/w) between -2 and 6.5, aqueous solubility (logS) between -6.5 and 0.5, predicted hERG liability (logHERG) < -5, number of predicted metabolic reactions (#metab) between 1 and 8, and no more than one violation of Lipinski's Rule of Five.Following application of these filters, a total of 65,998 compounds were retained across both libraries. These compounds exhibited favorable distributions of drug-like properties, as summarized in **Table 1**. The average molecular weight was 366.7 Da, with a logP mean of 3.39 and logS mean of -4.59, indicating moderate lipophilicity and acceptable solubility. Most compounds fell within the acceptable range for metabolic stability (mean #metab = 3.64) and had minimal Rule of Five violations (mean = 0.15). This refined set served as the basis for downstream docking, model training, and conditional generation using the JTNNVAE framework.

**Table 1.** *ADMET-based physicochemical profiling of the screened compound derived from Vitas and ChEMBL*.

| Property | Count | Mean | Std Dev | Min | 25% | 50% | 75% | Max |
|---|---|---|---|---|---|---|---|---|
| **MW** | 65998 | 366.68 | 70.31 | 142.24 | 316.42 | 361.35 | 411.45 | 668.72 |
| **logPo/w** | 65998 | 3.39 | 1.29 | -1.89 | 2.60 | 3.54 | 4.31 | 6.49 |
| **logS** | 65998 | -4.58 | 1.29 | -6.5 | -5.59 | -4.77 | -3.83 | 0.47 |
| **#metab** | 65998 | 3.64 | 1.76 | 1 | 2 | 3 | 5 | 8 |
| **Rule-Of-Five** | 65998 | 0.14 | 0.35 | 0 | 0 | 0 | 0 | 1 |

**3.3 Docking Analysis of Ternary Complexes Formed by E3 Ligases and Amyloid Beta-42 via Molecular Glue Compounds**

To investigate the potential of molecular glue compounds in promoting ternary complex formation between Aβ42 and E3 ligases, we performed molecular docking studies involving three key E3 ligases: VHL, CRBN, and MDM2. A representative small molecule with glue-like properties was docked to evaluate its ability to simultaneously interact with both Aβ42 and each of the E3 ligases, thereby stabilizing the formation of a ternary complex. **Figure 4a-c** panels display the structural configuration of the ternary complexes and the interaction highlights for each E3 ligase. Specifically, the protein structures of VHL, CRBN, and MDM2 are shown in complex with Aβ42 and the docked small molecule. The molecular glue bridges key residues on Aβ42 and the E3 ligases, forming a network of hydrogen bonds, hydrophobic contacts, and π-π stacking interactions essential for ternary complex stabilization.

The 2D interaction diagrams (**Figure 4d-f**) further delineate the nature of the binding interactions. In the VHL complex, key hydrogen bonds were observed between the molecular glue and residues such as GLU11, HIS13, and TYR112. For the CRBN complex, stabilizing interactions involved TRP86, HIS380, and ASN351. The MDM2 complex exhibited critical interactions with residues including GLU15, HIS19, and TRP23, contributing to the anchoring of the glue compound.

These docking results provide evidence for the potential of specific molecular glues to promote targeted protein degradation of Aβ42 by facilitating the formation of ternary complexes with E3 ligases. Although only one compound is shown here as a representative example, this analysis serves as a proof-of-concept for the broader applicability of molecular glue-based strategies in neurodegenerative disease intervention.

Following docking, molecular dynamics (MD) simulations were performed on the top-ranked ternary complexes formed by amyloid beta-42, the molecular glue compound, and each E3 ligase (VHL, CRBN, and MDM2). Only the highest-scoring docked compounds were selected for MD analysis to evaluate the stability and dynamic behavior of the ternary complexes over time. These simulations provided additional insights into the conformational stability and key intermolecular interactions sustaining the complex formation. As shown in **Supplementary Figure S3**, the RMSD trajectories of the VHL-ligand-Aβ42 (**S3a**), MDM2-ligand-Aβ42 (**S3b**), and CRBN-ligand-Aβ42 (**S3c**) complexes remain relatively stable throughout the simulation period, with no significant structural deviations. This suggests that the ternary assemblies are conformationally robust and maintain stable interactions over time. Consistent with this, hydrogen bond analysis further supports the persistence of ligand binding within the ternary interface. **Supplementary Figure S4** illustrates the time evolution of hydrogen bonds between the ligand and the E3 ligases in complex with Aβ42. The ligands consistently form multiple hydrogen bonds across the simulation, particularly in the VHL complex (**S4a**), indicating strong and sustained interaction networks. These findings reinforce the potential of our AI-generated compounds to facilitate stable ternary complex formation, a prerequisite for targeted Aβ42 degradation via the ubiquitin-proteasome system.

## 3.4 Training Data Compounds Characterization to Generate the Ligase-Conditioned Junction Tree Variational Autoencoder (LCJTVAE)

To develop LCJTVAE model capable of generating novel compounds that facilitate the E3 specific amyloid beta42 degradation. we first classified docking results into three affinity-based categories: High Affinity, Low Affinity, and No Affinity based on docking score thresholds less than -5kcal/mol, from -5 kcal/mol to -1.kcal/mol and above -1kcal/mol or positive values respectively as shown in **Figure 5**. **Table 2** shows that based on these docking score total of 65,998 compounds from both the library ChEMBL and Vitas, VHL had the highest number of high-affinity binders (e.g., 8,792 from ChEMBL and 6,607 from Vitas), while MDM2 exhibited the fewest. This stratification provided a robust and diverse dataset to train our AI model, ensuring that both positive (high/low affinity) and negative (no affinity) examples were well-represented.

**Table 2.** Summarizes the number of compounds falling into each affinity class for each ligase. A total of 65,998 compounds were evaluated across all conditions.

| Ligase | Library | High_Affinity | Low_Affinity | No_Affinity | Total |
|---|---|---|---|---|---|
| CRBN | ChEMBL | 4834 | 7448 | 36 | 12318 |
| CRBN | Vitas | 1748 | 11949 | 73 | 13770 |
| MDM2 | ChEMBL | 131 | 51 | 0 | 182 |
| MDM2 | Vitas | 2302 | 11361 | 86 | 13749 |
| VHL | ChEMBL | 8792 | 3414 | 80 | 12286 |
| VHL | Vitas | 6607 | 6949 | 137 | 13693 |
| Total | | | | | 65998 |

We also examined the correlation between docking scores and three key molecular descriptors: molecular weight (MW), lipophilicity (logPo/w), and aqueous solubility (logS) (**Figure 6**). In **Figure 6**, high-affinity compounds (blue) were predominantly clustered in the MW range of 300-500 Da, suggesting an optimal size window for ternary complex formation. For logPo/w, most high-affinity compounds fell within the range of 2-5, indicating that moderate lipophilicity may favor binding. In contrast, no-affinity compounds (green) (**Figure 6**) were more widely distributed and skewed toward higher solubility values (logS > -2), whereas high-affinity compounds generally had lower logS values (< -4), reflecting reduced aqueous solubility. These quantitative trends highlight that docking performance and binding potential are influenced by distinct ADMET property windows, which were integrated as features in the training dataset for the AI model.

To further dissect the structural characteristics associated with varying binding affinities, we conducted a scaffold analysis of compounds across the high, low, and no-affinity categories for each E3 ligase (VHL, CRBN, and MDM2) (**Figure 7**). Distinct trends emerged for each ligase, suggesting that specific structural motifs are favored depending on the binding partner.

For VHL (**Figure 7a**), high-affinity compounds predominantly contained substituted aromatic rings and flexible linkers incorporating nitrogen atoms, while low-affinity compounds were structurally simpler, often lacking key heteroatoms. No-affinity compounds tended to feature bulky fused-ring systems and sulfur-containing groups, which may contribute to steric hindrance or poor complementarity at the binding site.

In the case of MDM2 (**Figure 7b**), high-affinity scaffolds were characterized by biphenyl motifs and polar side chains, likely enhancing both hydrophobic packing and hydrogen bonding. In contrast, no-affinity scaffolds were bulkier and more rigid, often featuring heterocycles and extended conjugation.

For CRBN (**Figure 7c**), high-affinity compounds commonly included ether-linked alkyl chains and small heterocycles, suggesting that flexibility and size may be important determinants of binding. No-affinity compounds frequently exhibited complex fused polyaromatic systems and sulfur-containing linkers, which may disrupt favorable binding orientations. This scaffold-based categorization not only validates the docking-based affinity predictions but also highlights recurring ligase-specific chemical features, which are essential for guiding structure-aware AI model training and scaffold-based virtual screening efforts.

### 3.5 Graph and Junction Tree Representations of Molecules for Model Training

To facilitate the generation of chemically valid and interpretable molecular structures, we utilized dual representations for each molecule: a molecular graph and a junction tree, serving as structured inputs to the Ligase-Conditioned JT-VAE framework.

Each molecule was decomposed into a set of cliques-maximal subgraphs such as aromatic rings or functional groups-to capture its structural hierarchy (**Supplementary Figure S5a**). For example, in one representative compound, a brominated aromatic ring (black-circled) and a hydroxyl-substituted phenyl group (red-circled) were identified as distinct cliques. These cliques were assembled into a junction tree structure, where each node represents a clique and edges represent direct atom-sharing connections between cliques (**Supplementary Figure S5b**). In this example, node 14 corresponds to the brominated ring and node 15 to the hydroxyl-phenyl group, with node indices matching the entries in Supplementary **Table S1**, which lists the atom indices defining each clique.

During molecule generation, the model first predicts the junction tree scaffold to ensure chemical validity, followed by assembly of the full molecular graph based on atomic connectivity between cliques. The molecular graph representation (**Supplementary Figure S5c**) captures atomic-level detail, with nodes representing atoms and edges representing bonds; standard bonds are shown in blue, while bonds participating in rings are highlighted in red. This detailed graph is used in the assembly phase to accurately reconstruct chemically coherent structures. A complete listing of all cliques and corresponding edges for the training compounds (up to 50 molecules) is provided in Supplementary **Table S2**.

### 3.6 The Loss Function

The ligase-conditioned JTNNVAE model was trained for 500 epochs using the Adam optimizer with an initial learning rate of 1e-3 and an exponential learning rate decay (decay rate = 0.9 per step). To stabilize the training dynamics, all bias parameters were initialized to zero, and non-bias parameters were initialized using Xavier normal initialization. A KL annealing schedule was applied during training, with the KL divergence coefficient ($\beta$) linearly increased from 0 to 1 at a rate of 0.002 per batch.

To prevent gradient explosion, gradient clipping was applied with a max norm of 50. The total number of model parameters was approximately 5393K. The training loss, plotted in **Figure 8**, shows a sharp decline within the first 100-150 epochs, followed by a gradual convergence toward lower values, indicating stable model optimization.

### 3.7 AI-Generated Compounds and Model Performance
### 3.7.1 Molecular Diversity Visualization via t-SNE

To assess the chemical diversity and distribution of AI-generated compounds relative to training molecules, we employed t-distributed stochastic neighbor embedding (t-SNE) to project high-dimensional molecular fingerprints into two-dimensional space. To enable clear visualization of the chemical space, we randomly selected 100 training compounds for each E3 ligase (VHL, CRBN, and MDM2). Given that the number of AI-generated compounds was relatively small (up to 10 maxmimum), using the full training dataset-which is much larger-would have obscured the spatial distribution and diversity of the generated molecules. **Figure 9** displays t-SNE plots for VHL, MDM2, and CRBN-specific compound sets, visualizing both training compounds (blue) and AI-generated molecules (orange) in their respective embedding spaces.

For VHL (**Figure 9(a)**), the AI-generated compounds clustered tightly in a distinct region of the chemical space, yet remained within the broader distribution of the training molecules. This suggests that the model effectively captured scaffold-level diversity while preserving chemical validity. In the case of MDM2 (**Figure 9(b)**), generated compounds exhibited partial overlap with the training distribution, indicating moderate exploration of novel chemical space while still retaining features relevant to the target ligase. For CRBN (**Figure 9(c)**), the generated compounds formed a dense and centrally located cluster, reflecting the model's focus on optimizing around known chemical space with a bias toward validity and synthesizability.

Collectively, these visualizations demonstrate that the Ligase-Conditioned JT-VAE framework is capable of generating structurally diverse and target-specific molecular glues, while modulating novelty depending on training data distribution and target binding site complexity

### 3.7.2 Model Performance Across E3 Ligases

To evaluate the quality of molecules generated by our Ligase-Conditioned JT-VAE (LC-JT-VAE) framework, we assessed five key performance metrics: validity, uniqueness, novelty, average quantitative estimate of drug-likeness (QED), and Lipinski's rule-of-five adherence across three target E3 ligases: VHL, MDM2, and CRBN (**Figure 10**). All models consistently produced highly valid molecules, with validity scores exceeding 96%, confirming the chemical feasibility of the generated SMILES strings. The VHL model achieved the highest performance, with a validity of 96.30%, uniqueness of 85.19%, and novelty of 81.48%, indicating strong diversity and minimal overlap with the training set.

The CRBN-specific model generated molecules with 100% validity, although with slightly lower uniqueness (82%) and novelty (80%), suggesting a higher overlap with learned structural motifs. The MDM2 model showed balanced performance with 97.87% validity, 97.87% uniqueness, and 93.62% novelty, demonstrating strong generalization beyond training data. However, QED scores across all models remained moderate (0.35-0.42), reflecting the potential need for post hoc optimization to enhance drug-likeness. Similarly, Lipinski's adherence ranged from 60-83%, with the VHL-targeted model outperforming CRBN and MDM2, suggesting a better balance between molecular complexity and pharmacokinetic profiles. These metrics validate the LC-JT-VAE model's capability to generate novel, chemically valid, and E3-selective molecular glues, with each E3-specific model learning distinct chemical priors reflective of its ligase binding site context.

### 3.7.3 E3 ligase Protein-Specific Structural Insights into AI-Generated Compounds

The generated compounds for VHL (**Figure 11(a)**) displayed diverse ring systems, aliphatic chains, and heteroatoms (e.g., nitrogen, oxygen) that are consistent with features found in known VHL binders. Notably, several molecules contained ester, amide, and cyclic amine functionalities, which are important pharmacophoric elements for engaging VHL binding pockets.

For MDM2 (**Figure 11(b)**), the model generated the molecules having scaffolds enriched with aromatic and polycyclic structures, some bearing fused heterocycles or alkyl substitutions that resemble known MDM2-p53 interaction inhibitors. The generated compounds often featured extended planar regions, suggesting favorable π-π stacking potential within the MDM2 binding cleft.

In the case of CRBN (**Figure 11(c)**), the generated compounds included many small fragment-like structures and thalidomide-inspired motifs such as phthalimide cores and flexible linkers. These molecules tended to be chemically simpler yet retained functional groups commonly associated with CRBN modulators, indicating the model's ability to learn and replicate CRBN-specific ligand patterns.

Together, these AI-generated molecular glues show ligand diversity tailored to the unique binding environments of VHL, MDM2, and CRBN, supporting the effectiveness of our protein-conditioned generation strategy.

### 3.7.4  3D Conformational Plausibility of Ligase-Specific Compounds

To assess the 3D conformational stability of the ligase-specific compounds generated by our model, we applied RDKit's ETKDG algorithm to independently generate multiple conformers for each molecule. Although the **Supplement Table S3, S4, S5** does not report absolute energy values, the use of RDKit's ETKDG algorithm-which is designed to generate low-energy, experimentally plausible conformations-combined with the consistently low RMSD values between independent conformers, supports the interpretation that the generated ligase-specific compounds adopt stable and chemically reasonable 3D geometries. The conformational robustness of each compound was evaluated by calculating the root-mean-square deviation (RMSD) between two independently generated conformers (**Supplement Table S3, S4, S5**). RMSD values were negligibly small across all compounds-ranging from 0 to approximately $1.3 \times 10^{-7}$ Å-even for highly flexible molecules such as VHL_Cmpd_1 (13 rotatable bonds), VHL_Cmpd_8 (12 rotatable bonds) (**Supplement Table S3),** and MDM2_Cmpd_6 (22 rotatable bonds) (**Supplement Table S5)**. These minimal deviations indicate that the generated molecules consistently adopt low-energy, chemically plausible conformations, further validating the inclusion of torsional angle features in our Ligase-Conditioned JT-VAE framework. This analysis supports the model's capacity to produce structurally reliable compounds suitable for downstream structure-based drug design. The Python script used for this evaluation is provided in the Supplementary section (**Figure S5**).

### 3.8 Target-Specific Docking Validates the Precision of AI-Generated Compounds

To evaluate the binding specificity and interaction profiles of our AI-generated compounds, we performed systematic molecular docking of **all** designed molecules against each of the three E3 ligase-Aβ42 complexes (VHL, CRBN, and MDM2). For each compound, docking was carried out not only against its intended E3 ligase target but also cross-docked into the other two ligases to assess potential off-target binding. **Figure S4** presents examples of the **top-ranked compounds** based on docking scores for each E3 ligase: **VHL_Cmpd_4** for VHL (docking score: -5.98 kcal/mol in **Figure 12**), **CRBN_Cmpd_3** for CRBN (-5.80 kcal/mol in **Figure 12**), and **MDM2_Cmpd_5** for MDM2 (-5.78 kcal/mol in **Figure 12**). Structural visualizations in **Figure 12** illustrate that the top most ranked docked compounds engage their respective E3 ligase binding interfaces through multiple key interactions, including hydrogen bonding, hydrophobic contacts, and π-π stacking with critical pocket residues. For instance, VHL_Cmpd_4 forms hydrogen bonds with HIS13 of Aβ42 and HIS110 of VHL, whereas CRBN_Cmpd_2 establishes extensive interactions with HIS355, TRP402, and surrounding residues within the CRBN binding pocket.

To quantify the binding selectivity of all generated compounds, a heatmap of docking scores across all ligand-ligase combinations is shown in **Figure 12**. Compounds generally demonstrated stronger binding (more negative docking scores) to their designed target ligase compared to off-target ligases. Specifically, the average docking score for VHL-specific compounds against VHL was -5.84 kcal/mol, compared to an average of -4.05 kcal/mol and -4.15 kcal/mol when docked into MDM2 and CRBN, respectively. Similarly, CRBN-specific compounds exhibited an average docking score of -5.76 kcal/mol against CRBN, whereas the corresponding averages against VHL and MDM2 were markedly less favorable. These quantitative results further confirm that our LC-JT-VAE platform successfully generated compounds with strong binding affinity and target selectivity toward their intended E3 ligase.

## 4. DISCUSSION

In this study, we developed a **structure-guided generative framework** to design **E3 ligase-specific molecular glues** targeting **amyloid-β42 (Aβ42) fibrils**. Starting with the identification of a ligandable site on the Aβ42 fibril surface (**Figure 3**), we screened a large chemical space to discover compounds capable of bridging Aβ42 with key E3 ligases- **CRBN**, **VHL**, and **MDM2**. Following rigorous physicochemical and pharmacokinetic filtering, candidate molecules were evaluated for their ability to form ternary complexes through interaction modeling with each E3 ligase and Aβ42 **(Table 2)**. These prioritized compounds were subsequently used to train a **target-specific generative model** that integrates both the structural features of ligase binding sites and the 3D conformational dynamics of small molecules, incorporating torsional angles as an additional feature to enhance chemical realism.

Our results demonstrate that the generative model effectively captured **ligase-specific chemical preferences**, producing valid, novel, and structurally diverse compounds optimized for each E3 ligase (**Figure 9 and 10**). Performance evaluation showed consistently high chemical validity and target selectivity across generated compound sets (**Figure 10 and 12**). Dimensionality reduction analysis confirmed that the model generated molecules covering both novel and known chemical spaces, while detailed compound visualizations revealed diversity and preservation of critical pharmacophores associated with ligase engagement (**Figure 11**). These outcomes highlight the model's potential utility for designing degrader compounds against **structurally challenging targets** such as Aβ42.

The **comprehensive docking analysis** further underscores the effectiveness of our **LC-JT-VAE generative framework** in designing compounds with **targeted binding specificity**. By docking all generated compounds across multiple E3 ligases, we demonstrated that the molecules preferentially bind to their intended ligase targets rather than interacting nonspecifically across ligases. This specificity is crucial for the success of molecular glue-mediated ternary complex formation.

The **top-ranked compounds**, presented in **Figure S4**, provide illustrative examples of how our designed molecules establish **favorable interaction networks** within the ligase binding pockets, engaging critical residues through hydrogen bonding, hydrophobic contacts, and π-π interactions-key mechanisms for stabilizing ternary complexes.

Moreover, the **cross-target docking heatmap** (**Figure 12**) quantitatively supports this finding, showing that compounds exhibit **significantly stronger binding** (i.e., more negative docking scores) to their respective intended E3 ligases compared to off-target ligases. These results collectively validate the robustness of our AI-driven design platform, confirming its ability to generate chemically valid, structurally optimized, and functionally targeted therapeutic candidates.

Together, these findings position our integrated predictive and generative pipeline as a **powerful strategy** for **early intervention** in age-related diseases such as Alzheimer's disease, and offer a **generalizable framework** for therapeutic discovery against other undruggable targets.

## 5. CONCLUSION

This study presents an AI-driven framework for the rational design of E3 ligase-specific molecular glues targeting Aβ42, a key pathogenic peptide in Alzheimer's disease. By integrating ligand- and structure-based screening with advanced generative modeling, we demonstrate that the Ligase-Conditioned JT-VAE (LC-JT-VAE) can generate novel, drug-like compounds tailored for CRBN, VHL, and MDM2 ligases. These compounds exhibit strong potential to facilitate ternary complex formation and promote targeted degradation of intracellular Aβ42. The integration of protein sequence embeddings and torsional flexibility into the generative model significantly improves target specificity and chemical realism. Our approach sets the stage for AI-enabled discovery of therapeutic degraders against aggregation-prone proteins, offering new avenues in the treatment of Alzheimer's and other neurodegenerative disorders.

**Supplementary Materials**

The following supporting information can be downloaded at: https://github.com/thom-DEC/LCJTVAE-SupplementalData. Supplementary Data S1 is available.

**Author Contributions**

Conceptualization, N.N.I. and T.R.C.; methodology, N.N.I. and T.R.C.; software, N.N.I. and T.R.C.; validation, N.N.I. and T.R.C.; formal analysis, N.N.I. and T.R.C.; investigation, N.N.I. and T.R.C.; resources, N.N.I. and T.R.C.; data curation, N.N.I. and T.R.C.; writing-N.N.I. and T.R.C.; visualization, N.N.I.; supervision, T.R.C.; project administration, T.R.C.; funding acquisition, T.R.C. All authors have read and agreed to the published version of the manuscript.

**Funding**

T.R.C. was supported in part by the Center for Individualized Medicine (CIM) and the Florida Department of Health-Ed and Ethel Moore Alzheimer's Disease Research Program [24A19].

**Institutional Review Board Statement**

Not applicable.

**Informed Consent Statement**

Not applicable.

**Data Availability Statement**

Data and models will be made available upon request.

**Conflicts of Interest**

N.N.I and T.R.C. declare they have no competing interests. The funders had no role in the design of the study; in the collection, analyses, or interpretation of data; in the writing of the manuscript; or in the decision to publish the results.

**FIGURES.**

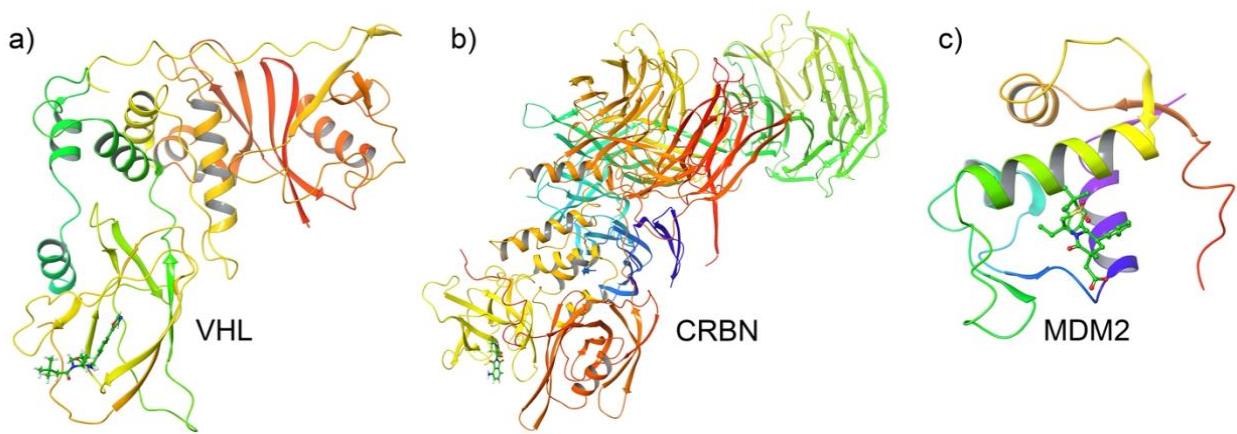

**Figure 1**. **Prepared 3D structures of representative E3 ligases used in this study, highlighting bound ligands in green stick representation**. (a) von Hippel-Lindau (VHL), (b) Cereblon (CRBN), and (c) Mouse Double Minute 2 homolog (MDM2). Each structure is visualized with ligands occupying their respective binding pockets, confirming appropriate docking orientation and active site integrity.

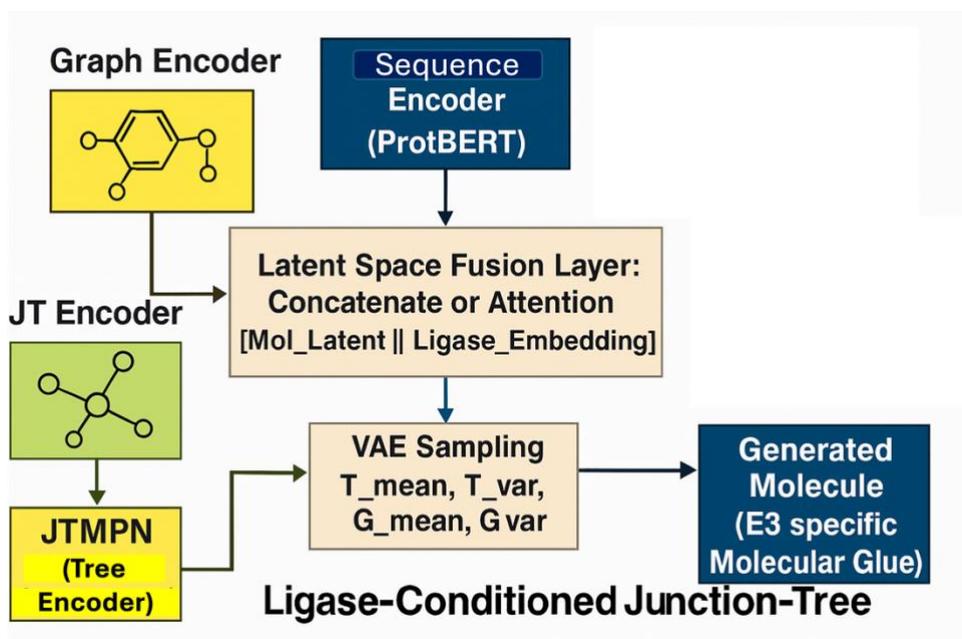

**Figure 2. Conditional JTNNVAE architecture for generating E3 ligase-specific molecular glues.** Molecular graphs are encoded via MPN, and junction trees via JTMPN, then passed to the JTNNEncoder. Simultaneously, binding site sequences from E3 ligases (e.g., CRBN, VHL, MDM2) are encoded into vector representations. These embeddings are fused to form a conditioned latent space, which is sampled using VAE parameters (T_mean, T_var, G_mean, G_var). The conditional JTNNDecoder reconstructs a chemically valid molecule tailored to the specific ligase context.

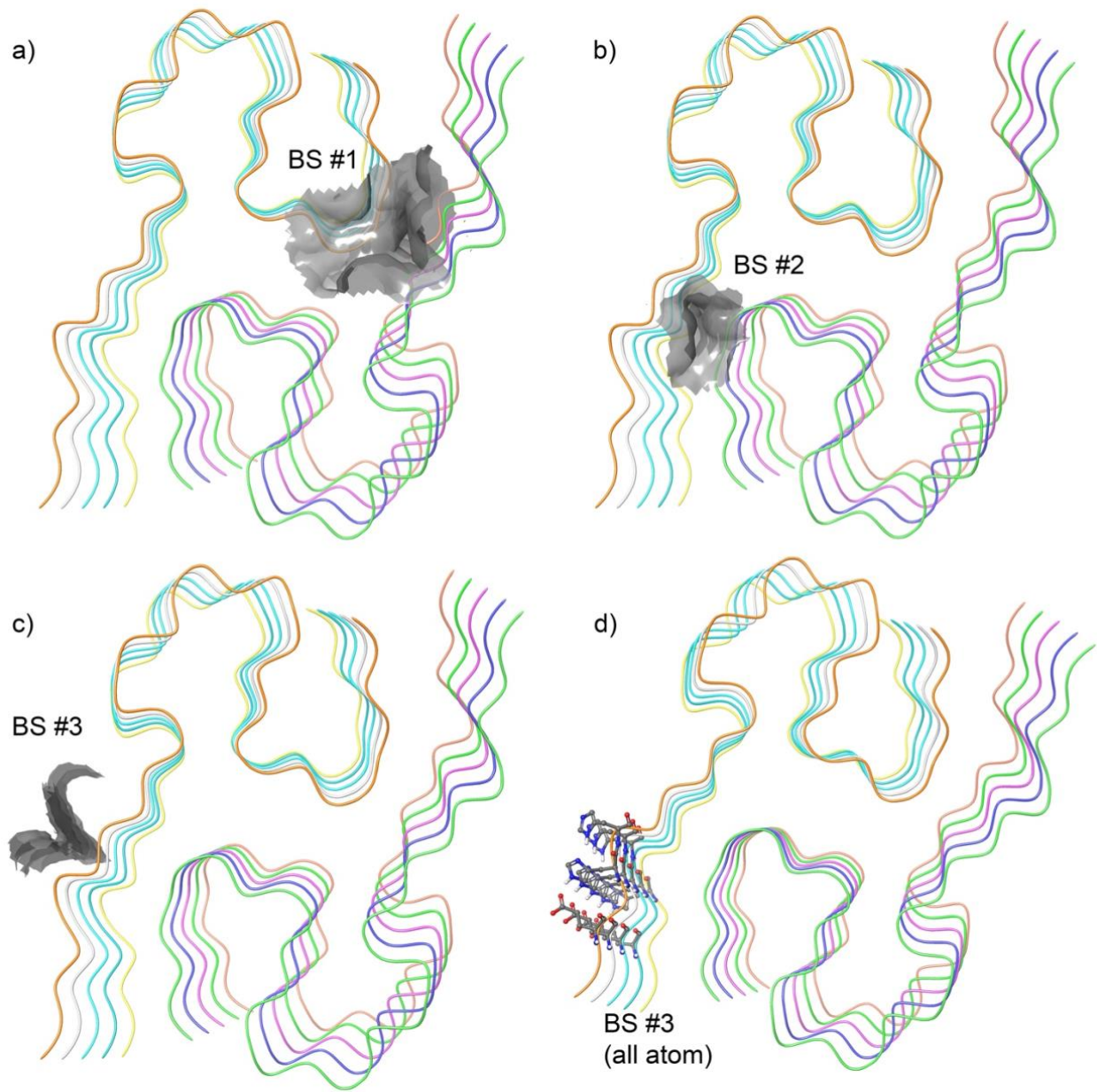

**Figure 3**. **Predicted binding sites (BS) on the amyloid beta 42 (Aβ42) peptide using Schrödinger SiteMap**. (a-c) Top three ranked potential binding pockets identified on the Aβ42 structure, with surface representations highlighting pocket volumes and spatial locations. Among these, site (c) was selected for subsequent molecular docking studies based on druggability and spatial characteristics. (d) Atomic-level representation of the selected binding pocket, showing detailed residue arrangement and pocket architecture to guide structure-based ligand design.

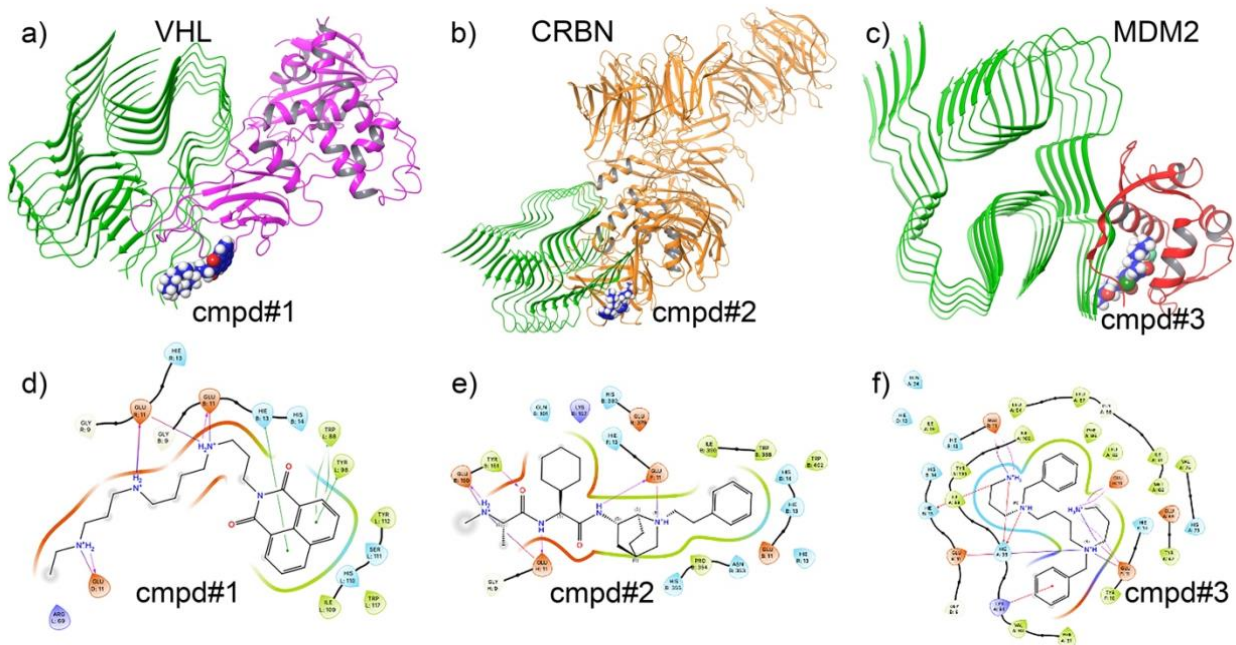

**Figure 4.** *Ternary complex formation between amyloid beta-42 (Aβ42), a representative molecular glue compound, and E3 ligases (VHL, CRBN, and MDM2)*. (a-c) 3D docking structures showing the formation of ternary complexes for (a) VHL-Aβ42-molecular glue, (b) CRBN-Aβ42-molecular glue, and (c) MDM2-Aβ42-molecular glue. The small molecule compound (cmpd) is shown in ball-and-stick representation, bound at the interface of Aβ42 and each E3 ligase. (d-f) 2D interaction diagrams corresponding to the docking poses for (d) VHL, (e) CRBN, and (f) MDM2 complexes. Intermolecular interactions include hydrogen bonds (purple lines), salt bridges, hydrophobic contacts (green arcs), and π-π stacking, indicating strong binding affinity and ternary complex stabilization.

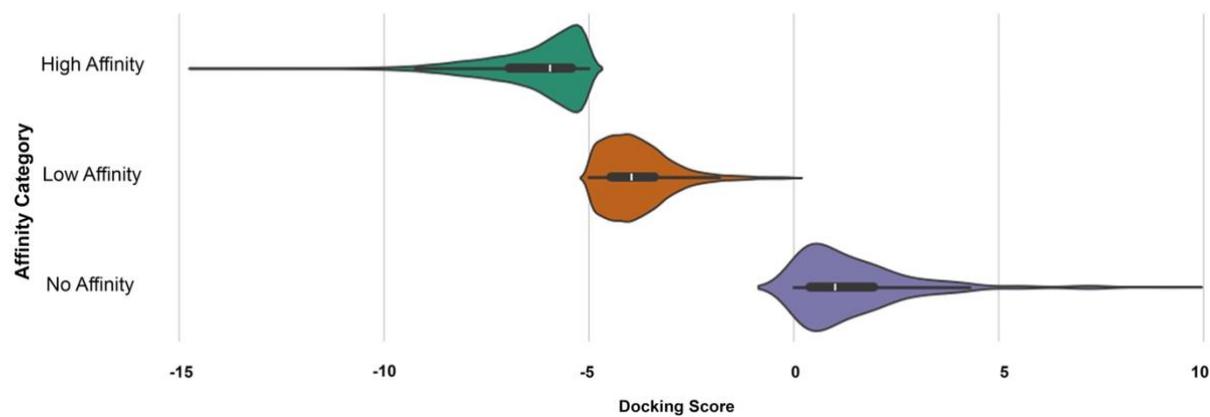

**Figure 5. Distribution of docking scores for compounds screened against E3 ligases VHL, CRBN, and MDM2, categorized by affinity class**. Docking scores are grouped into three affinity categories: High Affinity (≤ -5 kcal/mol), Low Affinity (-5 to -1kcal/mol), and No Affinity (≥ -1 kcal/mol).

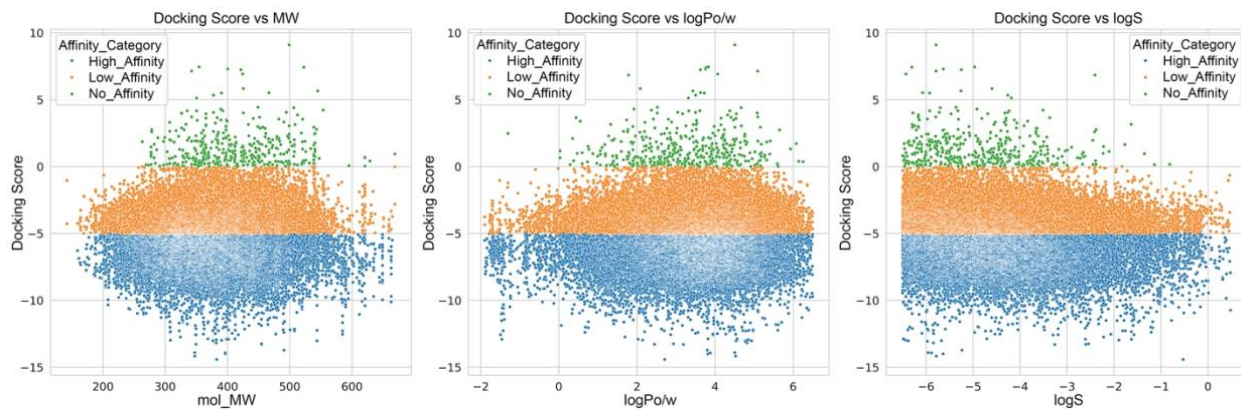

**Figure 6**. **Relationship between docking scores and selected ADMET properties across all screened compounds**. Compounds are color-coded by affinity category: High Affinity (blue), Low Affinity (orange), and No Affinity (green). (a) Docking scores vs. molecular weight (MW); (b) Docking scores vs. predicted aqueous solubility (logS); (c) Docking scores vs. predicted lipophilicity (logP). The plots illustrate the spread of affinity categories across diverse physicochemical properties, revealing that high-affinity compounds tend to cluster within drug-like regions, supporting their potential suitability for further development.

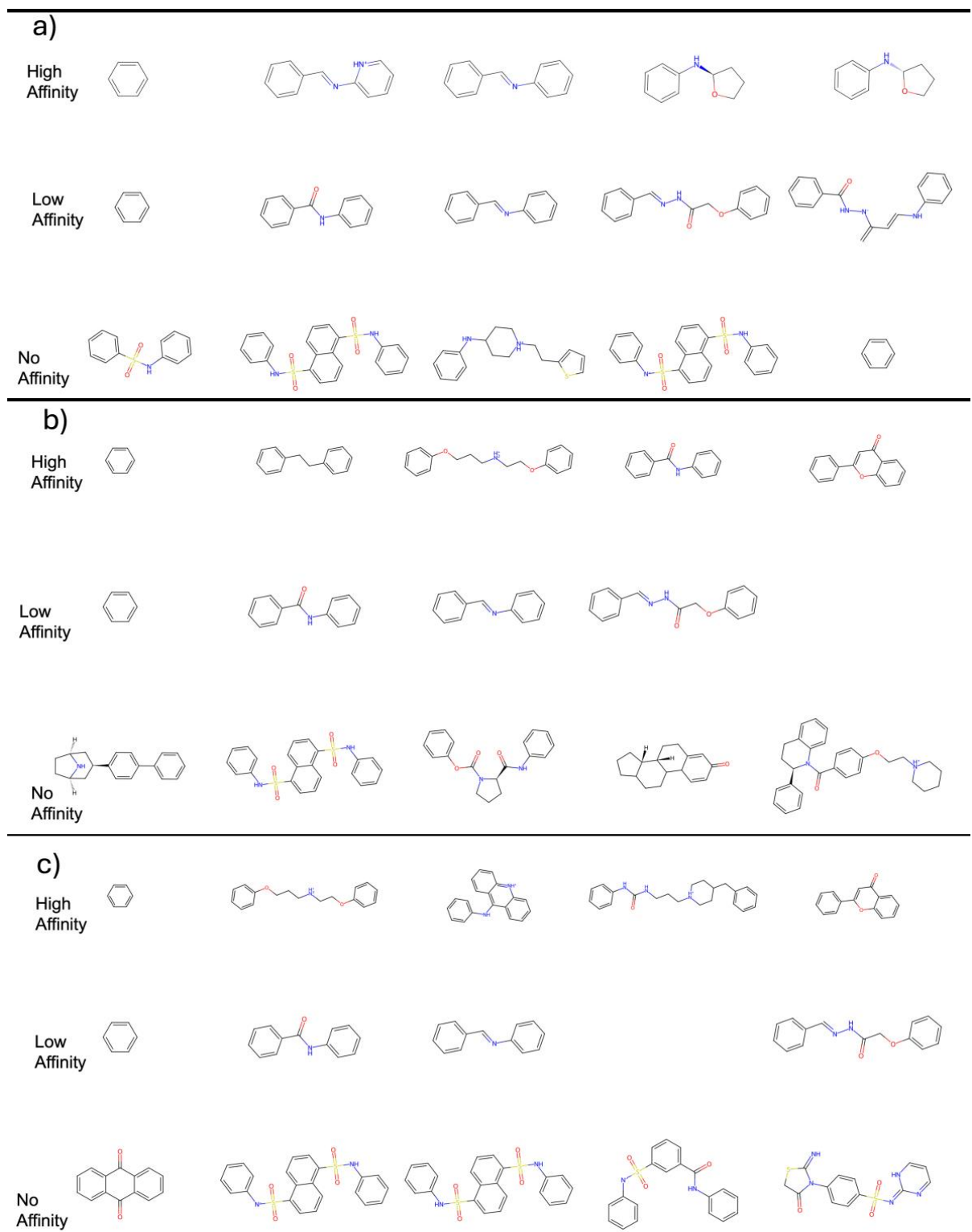

**Figure 7**. **Most frequently occurring molecular scaffolds identified across three affinity categories**-High Affinity (top), Low Affinity (middle), and No Affinity (bottom)-for each E3 ligase: (a) VHL, (b)MDM2, and (c) CRBN.

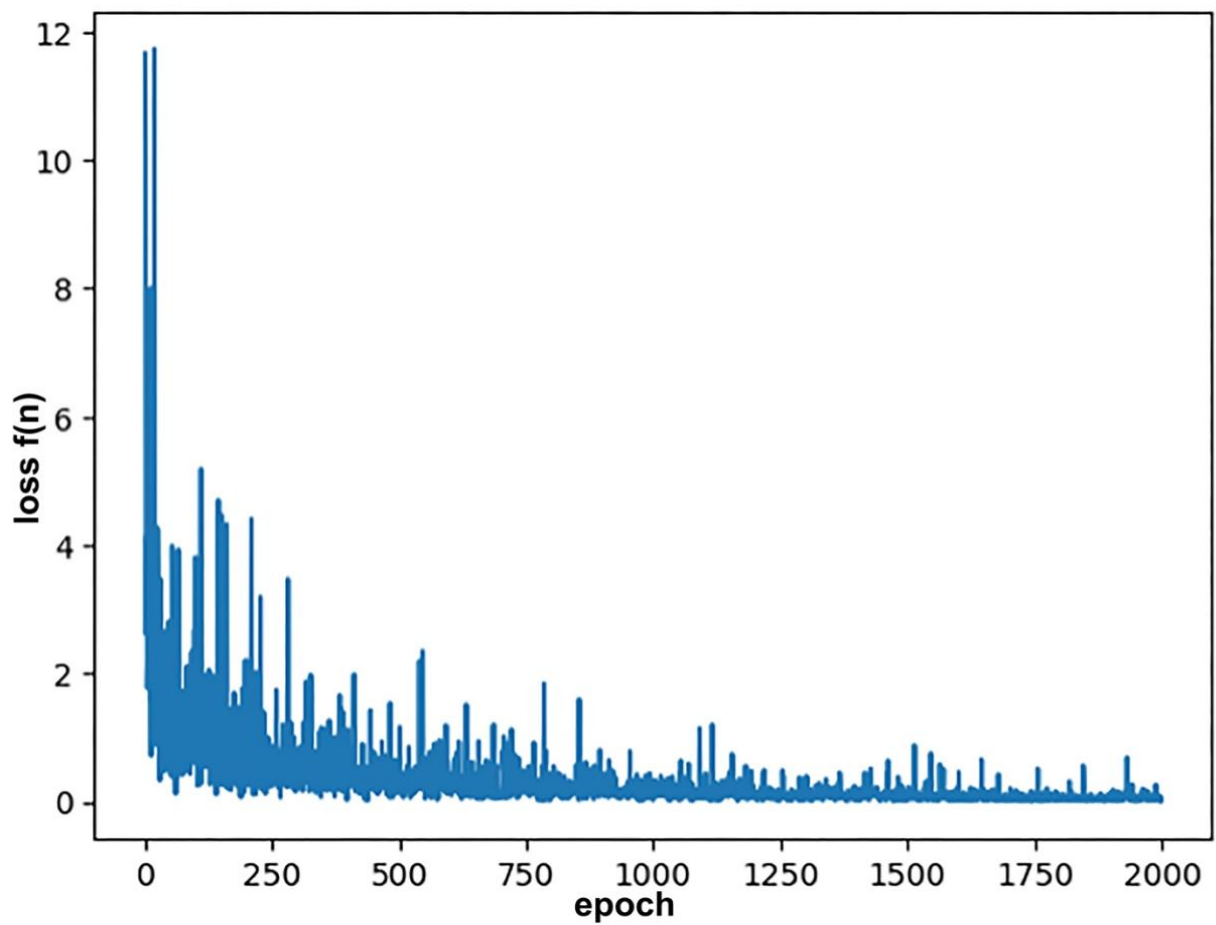

**Figure 8**. **Training loss curve of the JTNNVAE model trained on docked compounds targeting Amyloid beta42**. The sharp initial drop followed by gradual convergence indicates successful learning of chemically valid structures relevant to E3 ligase binding.

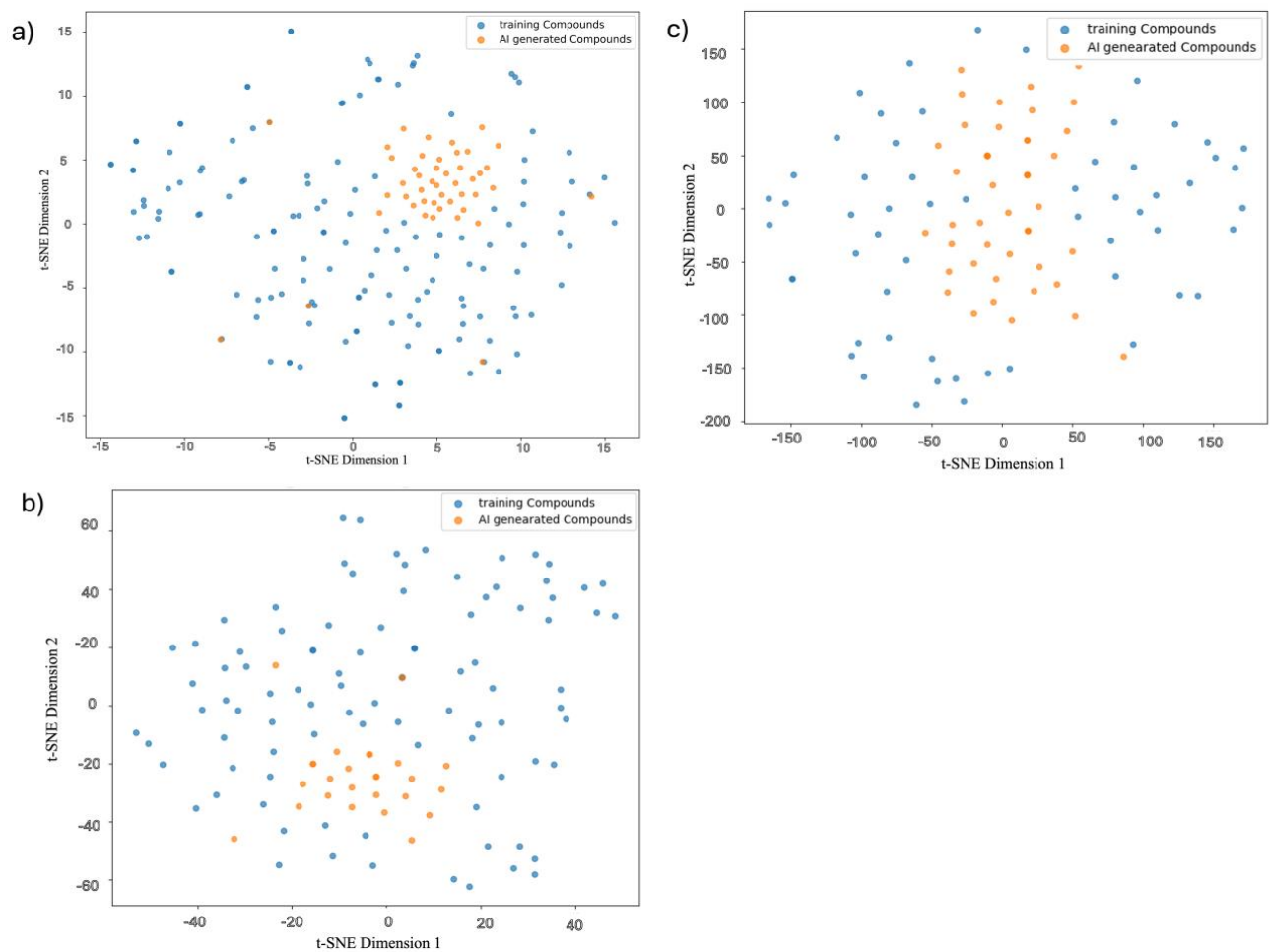

**Figure 9.** t-SNE plots visualizing the chemical space distribution of training compounds (blue) and AI-generated compounds (orange) for each E3 ligase: (a) VHL, (b) MDM2, and (c) CRBN.

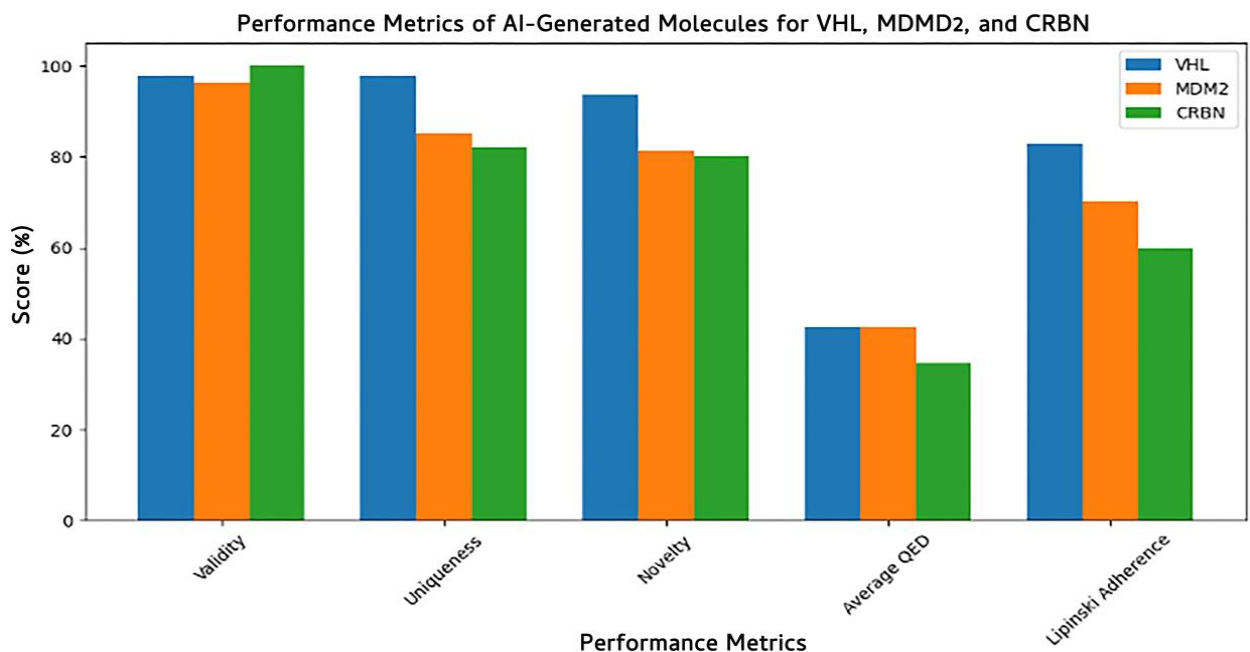

**Figure 10. Performance evaluation of AI-generated molecules for E3 ligases VHL, MDM2, and CRBN across key generative metrics**. Metrics include validity, uniqueness, and novelty of generated compounds, as well as average quantitative estimate of drug-likeness (QED) and compliance with Lipinski's Rule of Five. High validity and novelty scores indicate the model's ability to generate chemically sound and diverse structures, while QED and Lipinski adherence reflect drug-like potential of the designed molecules.

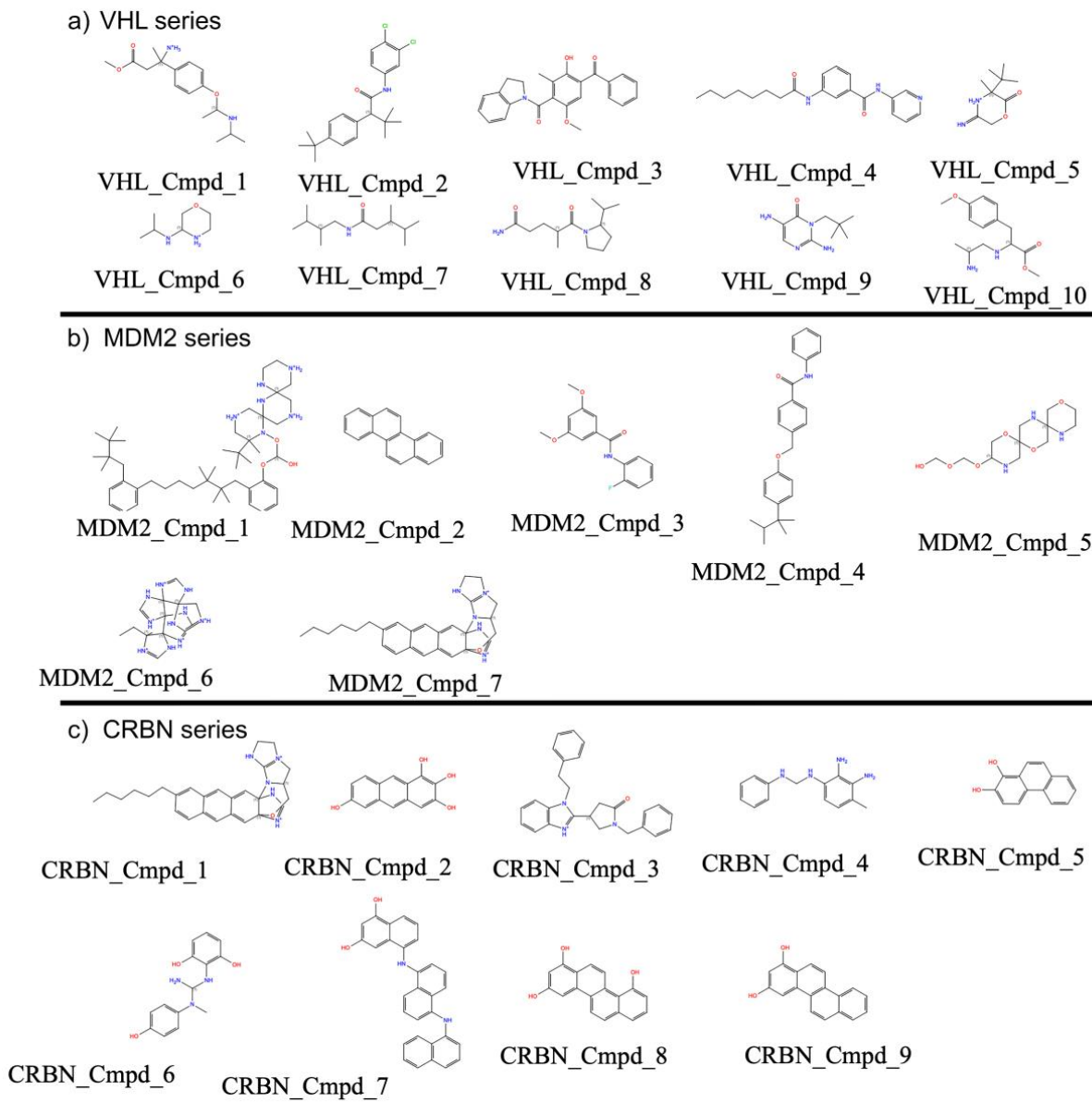

**Figure 11**. Representative AI-generated compounds designed for three E3 ligases: (a) VHL, (b) MDM2, and (c) CRBN. The series of exemplar compounds illustrates the distinct scaffolding preferences and functional group diversity observed from our results.

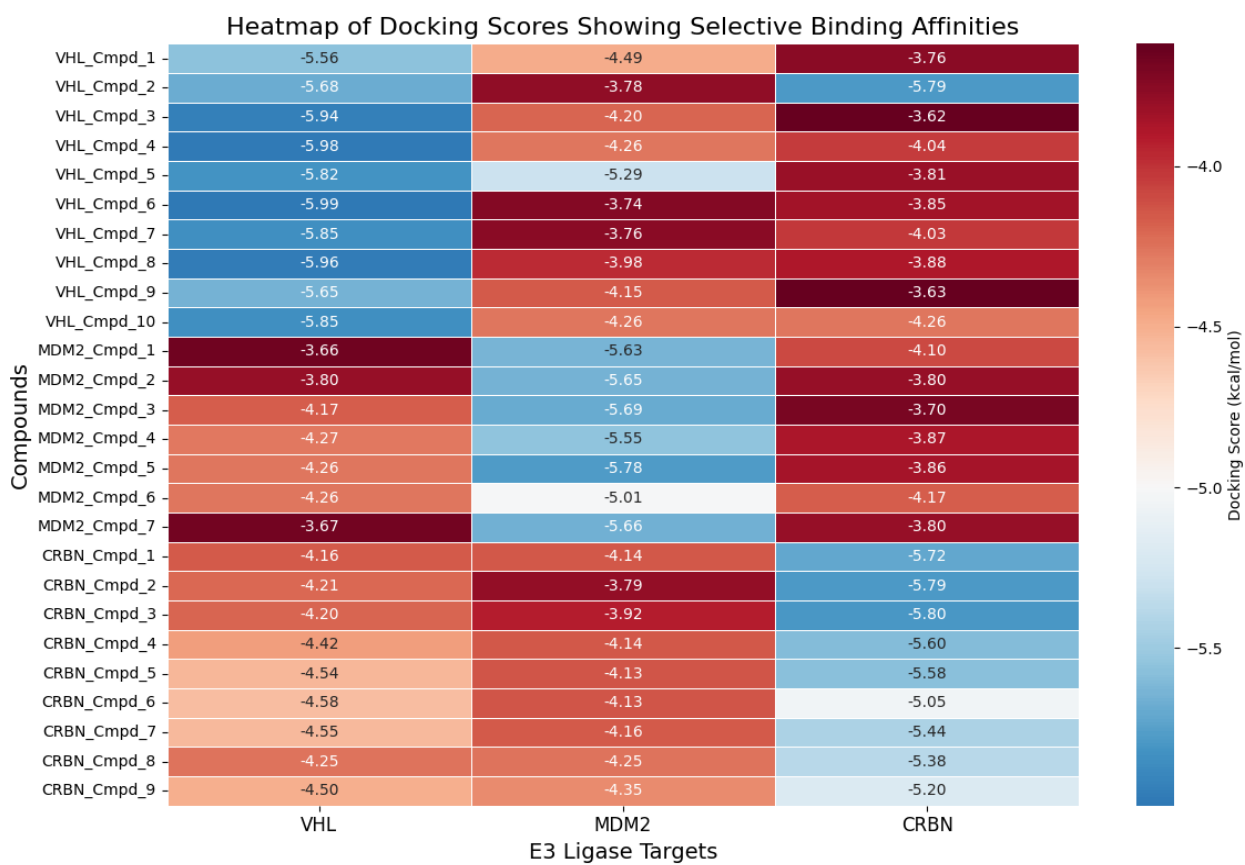

**Figure 12.** **Heatmap of docking scores (kcal/mol) for AI-generated compounds across VHL, CRBN, and MDM2 ligases**. Each row represents a compound, and each column corresponds to an E3 ligase. Stronger (more negative) binding affinities are shown in blue. The results demonstrate target-specific binding, with designed compounds showing preferential affinity for their intended ligases and lesser binding affinity for their unintended targets.